\documentclass[sigconf, screen]{acmart}
\AtBeginDocument{%
  }

\setcopyright{acmlicensed}
\copyrightyear{2025}
\acmYear{2025}
\acmDOI{XXXXXXX.XXXXXXX}
\acmConference[Conference acronym 'XX]{Make sure to enter the correct
  conference title from your rights confirmation email}{June 03--05,
  2025}{Woodstock, NY}
\acmISBN{978-1-4503-XXXX-X/2025/06}

\PassOptionsToPackage{table,xcdraw,HTML}{xcolor}
\usepackage{xcolor}
\usepackage{colortbl}
\usepackage{multirow}
\usepackage{pifont}

\begin{document}

\title{AlignGen: Boosting Personalized Image Generation with Cross-Modality Prior Alignment}

\author{Yiheng Lin}
\authornote{Both authors contributed equally to this research.}
\affiliation{
  \institution{Institute of Information Science, Beijing Jiaotong University}
  \country{China}
}
\affiliation{
  \institution{MT Lab, Meitu Inc.}
  \country{China}
}

\author{Shifang Zhao}
\authornotemark[1]  
\affiliation{
  \institution{Institute of Information Science, Beijing Jiaotong University}
  \country{China}
}

\author{Ting Liu}
\affiliation{
  \institution{MT Lab, Meitu Inc.}
  \country{China}
}

\author{Xiaochao Qu}
\affiliation{
  \institution{MT Lab, Meitu Inc.}
  \country{China}
}

\author{Luoqi Liu}
\affiliation{
  \institution{MT Lab, Meitu Inc.}
  \country{China}
}

\author{Yao Zhao}
\affiliation{
  \institution{Institute of Information Science, Beijing Jiaotong University}
  \country{China}
}

\author{Yunchao Wei}
\authornote{Corresponding author.}
\affiliation{
  \institution{Institute of Information Science, Beijing Jiaotong University}
  \country{China}
}

\renewcommand{\shortauthors}{Trovato et al.}

\begin{abstract}
Personalized image generation aims to integrate user-provided concepts into text-to-image models, enabling the generation of customized content based on a given prompt. Recent zero-shot approaches, particularly those leveraging diffusion transformers, incorporate reference image information through a multi-modal attention mechanism. This integration allows the generated output to be influenced by both the textual prior from the prompt and the visual prior from the reference image. However, we observe that when the prompt and reference image are misaligned, the generated results exhibit a stronger bias toward the textual prior, leading to a significant loss of reference content. To address this issue, we propose AlignGen, a Cross-Modality Prior \textbf{Align}ment mechanism that enhances personalized image generation by: 1) introducing a learnable token to bridge the gap between the textual and visual priors, 2) incorporating a robust training strategy to ensure proper prior alignment, and 3) employing a selective cross-modal attention mask within the multi-modal attention mechanism to further align the priors. Experimental results demonstrate that AlignGen outperforms existing zero-shot methods and even surpasses popular test-time optimization approaches.


\end{abstract}




\begin{teaserfigure}
  \includegraphics[width=\textwidth]{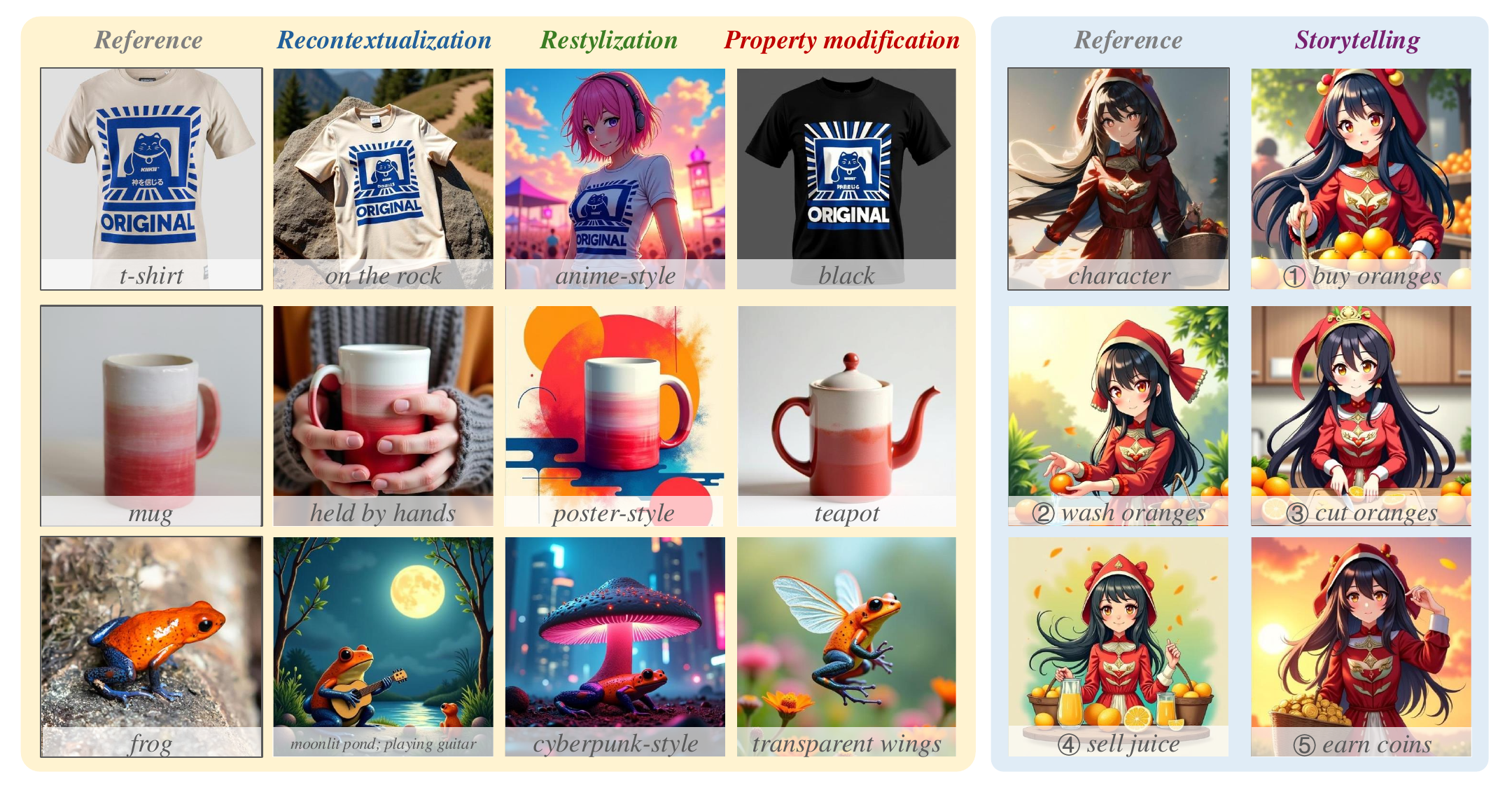}
  \caption{Given a reference image and prompt, our AlignGen model generates images that maintain consistent concepts and follow the prompt, without requiring test-time fine-tuning. AlignGen supports various applications, including recontextualization, restylization, property modification, and storytelling.}
  \Description{Given a reference image and prompt, our AlignGen model generates images that maintain consistent concepts and follow the prompt, without requiring test-time fine-tuning. AlignGen supports various applications, including recontextualization, restylization, property modification, and storytelling.}
  \label{fig:teaser}
\end{teaserfigure}


\maketitle

\section{Introduction}
\label{sec:introduction}
Recent advancements in image generation models, particularly diffusion models based on the diffusion transformer \cite{peebles2023scalable}, such as Stable Diffusion 3 \cite{esser2024scaling} and FLUX \cite{flux2024}, have demonstrated remarkable capabilities in generating high-quality images. These models are trained on billions of image-text pairs, achieving strong image-text alignment. However, effectively integrating user-provided concepts for personalized image generation remains crucial, as text alone may not fully capture user intent. The primary challenge is the generated image must follow the prompt description while preserving the details of the provided concept. 

To address this challenge, two primary approaches have emerged. First, test-time optimization techniques, such as Textual Inversion \cite{gal2023an} and DreamBooth \cite{ruiz2023dreambooth}, fine-tune pre-trained text-to-image models with a few reference images. While these methods excel at concept preservation, their need for extensive optimization steps introduces significant computational overhead. In contrast, zero-shot approaches eliminate test-time optimization by training image encoders on large-scale datasets, offering greater efficiency. Recent zero-shot research \cite{cai2024diffusion, tan2024ominicontrol} leverages the in-context generation capabilities of diffusion models to produce high-quality image pairs that feature the same reference concept in varying contexts. These methods utilize the built-in VAE encoder and DiT layers to encode the reference image into reference image tokens. These tokens are then combined with noisy image tokens and text tokens in the multi-modal attention mechanism.  However, despite utilizing such training data, current zero-shot methods still fall short of test-time optimization techniques in terms of preserving concept fidelity. 

\begin{figure}
    \centering
    \includegraphics[width=0.45\textwidth]{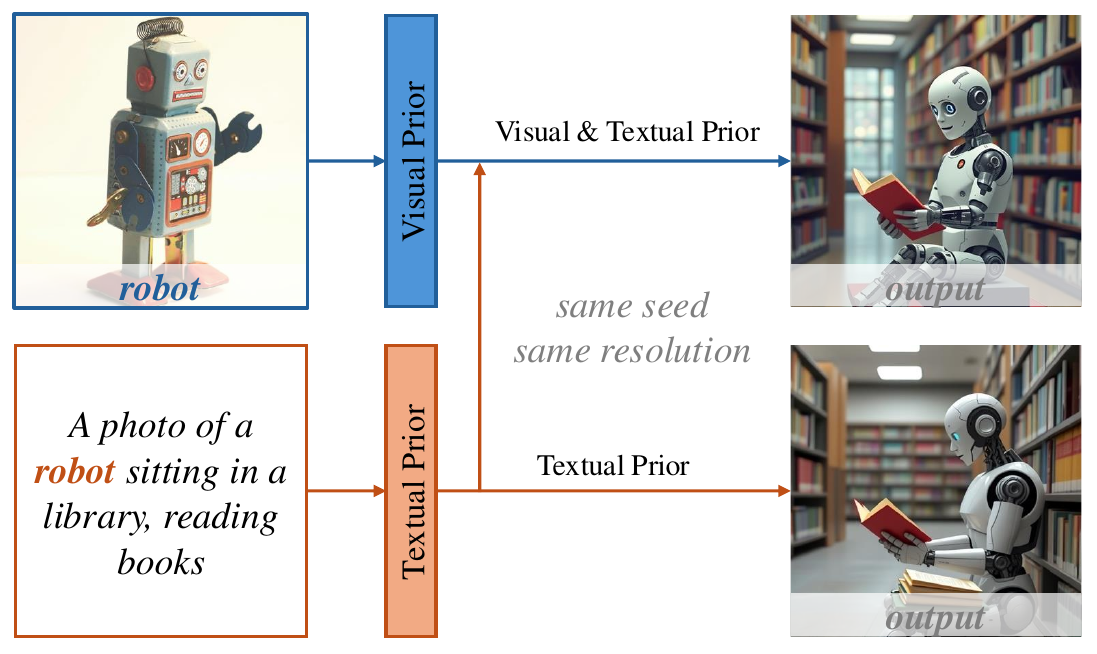}
    \caption{Explanation of Cross-Modality Prior Misalignment. The upper branch incorporates both the visual and textual priors in the multi-modal attention mechanism, while the lower branch integrates only the textual prior.}
    \Description{Explanation of Cross-Modality Prior Misalignment. The upper branch incorporates both the visual and textual priors in the multi-modal attention mechanism, while the lower branch integrates only the textual prior.}
    \label{fig:cross_modality_prior_mis}
\end{figure}

\textit{Why are zero-shot methods not as effective as test-time optimization?} Although multi-modal attention allows the generation to be guided by both the textual prior from the prompt and the visual prior from the reference image, \textbf{the generated results show a stronger bias toward the textual prior when the prompt and the reference image are misaligned.} As shown in Figure \ref{fig:cross_modality_prior_mis}, we generate images using the same prompt, seed, and resolution. In the top branch, both visual and textual priors are incorporated, while in the bottom branch, only the textual prior is used by replacing the reference image with a black image. Since text-to-image generation models are typically trained on large-scale datasets, the textual prior (e.g., "robot" in the prompt) often corresponds to a commonly occurring robot depiction (e.g., a humanoid robot), which is misaligned with the reference robot. The results indicate that both branches produce similar outputs, suggesting that the textual prior predominantly influences the generation process, leading to the loss of reference robot.

To mitigate this misalignment, we propose a cross-modality prior alignment mechanism. As outlined in Section \ref{sec:cross_modality_prior_alignment}, we introduce a learnable token before the concept words in the prompt and employ a deviation extraction module to enable the token to capture the deviation between textual and visual priors. Additionally, we apply a selective cross-modal attention mask to bind the concept word in the prompt with the visual prior in the reference image, further enhancing cross-modality alignment. During training, we randomly drop reference images and adjust concept words in the prompt, enabling the learnable token to effectively capture cross-modality deviations.

By aligning the textual and visual priors, our method demonstrates strong concept preservation and prompt following. As shown in Figure \ref{fig:teaser}, the user provides a reference image and a prompt for personalized image, and our method supports a variety of applications, including recontextualization (generating images of the given concepts in different environments with high preservation of the concept details and realistic scene-concept interactions), restylization (creating artistic renditions of a given concept), property modification (altering attributes like color and shape), and storytelling (generating different storyboards for anime characters). Furthermore, comprehensive experiments on DreamBench++ \cite{peng2024dreambench} demonstrate that our method achieves an optimal balance between concept preservation and prompt following, outperforming existing zero-shot methods and even surpassing test-time optimization approaches like DreamBooth LoRA. 

We summarize our contributions as follows:
\begin{itemize}
\item We identify the misalignment between cross-modal priors as a key factor in the loss of reference concepts.

\item We propose a Cross-Modality Prior Alignment mechanism that integrates a learnable token, deviation extraction module, training strategy, and selective cross-modal mask to align textual and visual priors, enhancing concept preservation ability.

\item Both qualitative and quantitative experiments on DreamBench++ demonstrate that our method achieves a better balance between concept preservation and prompt following than baselines.

\end{itemize}

\section{Related Work}
\subsection{Text-to-Image Generation}
Recent advancements \cite{esser2024scaling, flux2024, rombach2022high, nichol2022glide, ramesh2022hierarchical, saharia2022photorealistic} in diffusion models \cite{ho2020denoising} have significantly enhanced the field of text-to-image generation. Early models based on the U-Net \cite{ronneberger2015u} architecture, such as Stable Diffusion 1.5 \cite{rombach2022high} and SDXL \cite{rombach2022high}, utilize an autoencoder to map images into a latent space, where a U-Net model is employed for denoising. Recently, the diffusion transformer (DiT) \cite{peebles2023scalable} replaces the U-Net with a transformer architecture \cite{vaswani2017attention, dosovitskiy2021an}, demonstrating improvements in scalability. And the diffusion transformer with flow matching \cite{lipman2023flow} has become a dominant design paradigm, with models like Stable Diffusion 3 \cite{esser2024scaling} and FLUX \cite{flux2024} achieving remarkable performance improvements. Our method builds upon FLUX.1 Dev due to its robust image-text alignment and its open-source nature, making it an ideal foundation for our approach.

\subsection{Test-time Optimization Methods}

Test-time optimization methods \cite{han2023svdiff, voynov2023p+, alaluf2023neural, liu2023cones, avrahami2023break} aim to adapt pre-trained text-to-image models to personalized image generation models by using a limited number of user-provided examples. However, full fine-tuning of large models can be computationally expensive and may result in the loss of pre-trained universal knowledge. To mitigate this, Textual Inversion \cite{gal2023an} freezes the generative model and instead learns new "pseudo-words" in the text embedding space to represent novel concepts from a few images. DreamBooth \cite{ruiz2023dreambooth} fine-tunes the entire model to associate a unique identifier with a specific concept while preserving existing knowledge through a class-specific prior loss. DreamBooth LoRA trains an additional LoRA \cite{hu2022lora} instead of the entire model, which reduces the number of tuning parameters. Similarly, Custom Diffusion \cite{kumari2023multi} focuses on efficiently fine-tuning a small subset of the cross-attention layer parameters in diffusion models to embed new concepts. This approach allows for both single-concept personalization and the composition of multiple novel concepts. These methods aim to provide the ability to generate personalized content, but they come with significant training costs for each new concept.

\subsection{Zero-shot Methods}
Unlike test-time optimization methods, zero-shot methods \cite{wei2023elite, ye2023ip, li2023blip, ma2024subject, xiao2024fastcomposer, xiao2024omnigen} aim to provide an off-the-shelf ability to customize text-to-image models for given examples in a zero-shot manner, without any test-time training burden. BLIP-Diffusion \cite{li2023blip} utilizes BLIP-2 \cite{li2023blip2} to extract a text-aligned visual representation of the subject and injects it into the text encoder of a latent diffusion model (Stable Diffusion). IP-Adapter \cite{ye2023ip} presents a lightweight adapter that equips pre-trained text-to-image diffusion models with image prompt capabilities. Its key design is a decoupled cross-attention mechanism, which adds separate cross-attention layers for text and image features within the UNet of the diffusion model. Emu2 \cite{sun2024generative}, trained on vast multimodal sequence data with a unified autoregressive objective and an image encoder for input images, demonstrates strong multimodal in-context learning abilities for tasks requiring immediate reasoning, such as visual prompting and object-related generation. OminiControl \cite{tan2024ominicontrol} proposes a minimal and universal control method for DiT \cite{peebles2023scalable} architectures, handling both subject-driven generation and spatially-aligned tasks. It directly uses the generated model itself as the encoder, rather than an extra module, showing greater parameter efficiency. Diffusion Self-Distillation \cite{cai2024diffusion} employs a parallel processing architecture, treating the input image as the first frame of a video and generating a two-frame video as output for conditional editing. Although existing methods provide valuable advancements in reference image injection, they overlook the gap between reference images and prompt priors. By addressing this Cross-Modality Prior Misalignment, our method more efficiently leverages the reference image while maintaining concept preservation and prompt following capability.

\section{Methods}
\subsection{Preliminaries}

\begin{figure}
    \centering
    \includegraphics[width=0.45\textwidth]{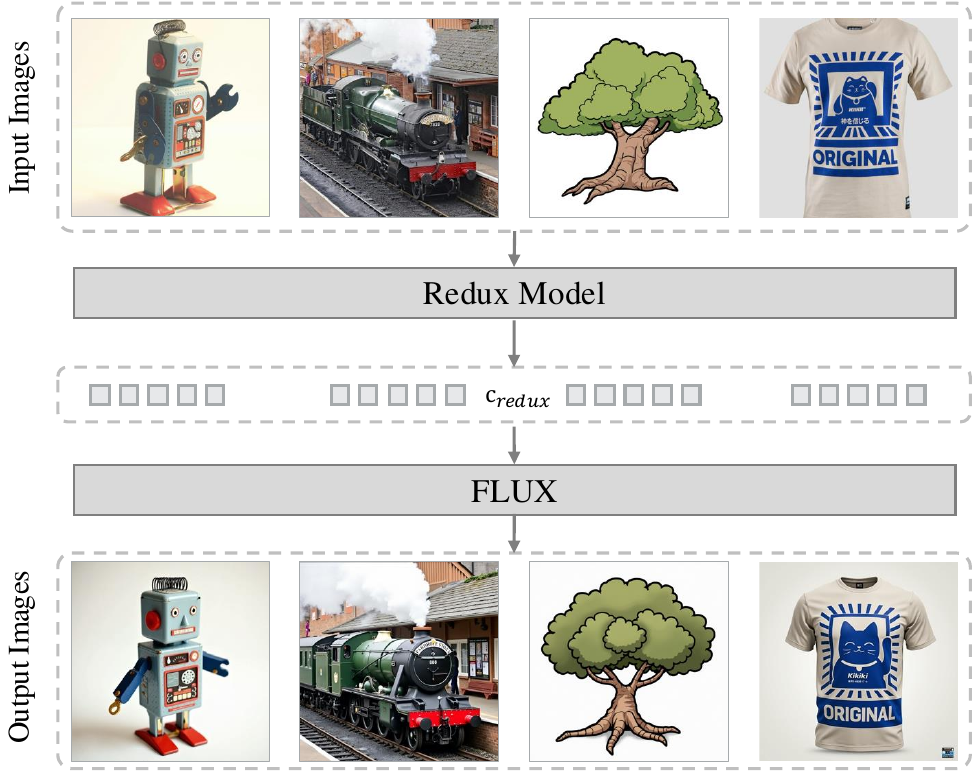}
    \caption{Visualization of reconstruction of input images by the Redux model. The prior encoded in the redux token ensures that the generated images preserve the color, shape, and style of the input, but it does not retain the fine details of the subjects.}
    \Description{Visualization of reconstruction of input images by the Redux model. The prior encoded in the redux token ensures that the generated images preserve the color, shape, and style of the input, but it does not retain the fine details of the subjects.}
    \label{fig:redux}
\end{figure}
\begin{figure*}
    \centering
    \includegraphics[width=0.98\textwidth]{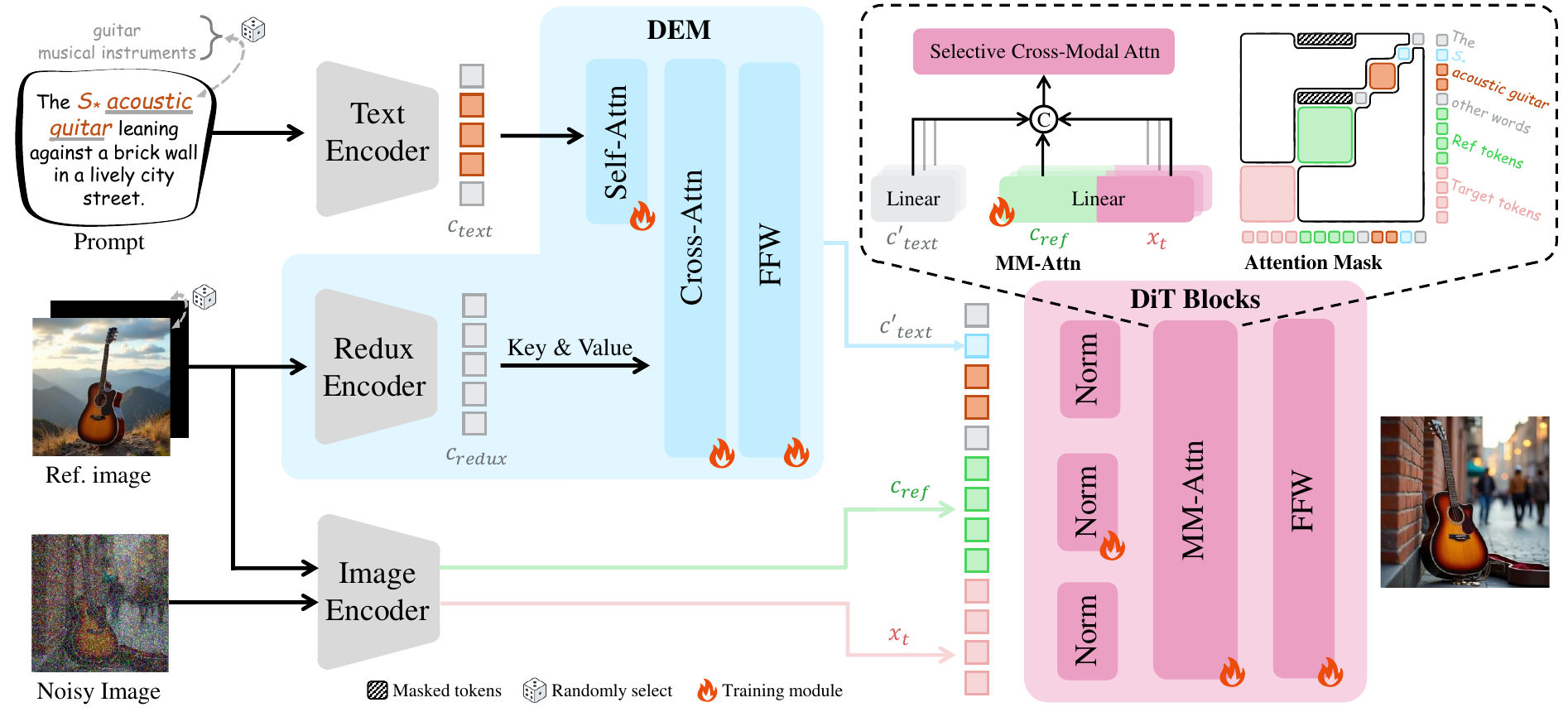}
    \caption{Overview of our pipeline. The prompt and reference image are first encoded into text tokens $c_{text}$ and redux tokens $c_{redux}$. The Deviation Extraction Module (DEM) then updates $c_{text}$ to $c_{text}'$, which becomes more aligned with the visual prior from the reference token $c_{ref}$. Finally, all tokens are concatenated and processed using multi-modal attention. Note that both the reference token and noisy image token share the same modules, with LoRA applied only to the reference token. Modules marked with a flame symbol are trainable, while the others remain frozen.}
    \Description{Overview of our pipeline. The prompt and reference image are first encoded into text tokens $c_{text}$ and redux tokens $c_{redux}$. The deviation extraction module (DEM) then updates $c_{text}$ to $c_{text}'$, which becomes more aligned with the visual prior from the reference token $c_{ref}$. Finally, all tokens are concatenated and processed using multi-modal attention. Note that both the reference token and noisy image token share the same modules, with LoRA applied only to the reference token. Modules marked with a flame symbol are trainable, while the others remain frozen.}
    \label{fig:framework}
\end{figure*}

\noindent\textbf{Diffusion Transformer.} Recent DiT models \cite{flux2024, rombach2022high} exhibit strong in-context generation capability, enabling the generation of grids within a single image where each grid maintains consistent attributes (e.g., subject identities, styles) while altering poses or layouts. This ability reflects the powerful text-image alignment capability of DiT models, which is driven by advanced text encoder \cite{raffel2020exploring} and multi-modal attention (MMA) mechanism. The text encoder accurately encodes prompts, while MMA facilitates full interaction between noisy image tokens and text condition tokens. Specifically, given noisy image tokens $x_t \in \mathbb{R}^{N \times d}$ and text condition tokens $c_{text} \in \mathbb{R}^{M \times d}$, where $M$ is the number of text tokens, $N$ is the number of noisy image tokens, $d$ is the dimension of each token, the MMA is computed as:
\begin{equation}
MMA([x_t; c_{text}]) = \text{Softmax} \left( \frac{Q K^\top}{\sqrt{d}} \right) V \in \mathbb{R}^{(M+N) \times d},
\end{equation}
where $[x_t; c_{text}]$ denotes the concatenation of image and text tokens, and $Q$, $K$, and $V$ represent the query, key and value corresponding to $[x_t; c_{text}]$.

Building on this, recent customized image generation techniques \cite{cai2024diffusion, tan2024ominicontrol} leverage the built-in VAE encoder and DiT Layers of FLUX \cite{flux2024} to encode the reference image into reference image tokens $c_{ref} \in \mathbb{R}^{N \times d}$. These reference tokens are then concatenated with $x_t$ and $c_{text}$ and processed via MMA:
\begin{equation}
MMA([x_t; c_{text}; c_{ref}]) = \text{Softmax} \left( \frac{Q' K'^\top}{\sqrt{d}} \right) V' \in \mathbb{R}^{(M+2N) \times d}.
\end{equation}
Although this approach integrates reference information with minimal architectural modifications, the generated results exhibit a stronger bias toward the textual prior when there is misalignment between $c_{text}$ and $c_{ref}$.

\noindent\textbf{Redux produces Text-Aligned Visual Representation.} 
The FLUX.1 Redux model \cite{flux2024} serves as an adapter for all FLUX.1 base models, designed for image variation generation. As illustrated in Figure \ref{fig:redux}, the Redux model encodes an input image into Redux tokens, denoted as $c_{redux} \in \mathbb{R}^{729 \times d}$. These tokens are then used as prompt embeddings to enable FLUX.1 to generate images with similar color and shape, demonstrating that the redux tokens contain rich visual representations of the reference image. Furthermore, redux tokens share the same positional index $(0,0)$ as text tokens $c_{text}$ in the RoPE \cite{su2024roformer} mechanism and utilize the same normalization layers, query/key/value projections, and MLP projections within the DiT blocks. This consistency suggests that redux tokens reside in a shared representation space with text tokens. Although the redux model can generate a text-aligned visual representation of the reference image, the redux tokens $c_{redux}$ cannot replace the reference tokens $c_{ref}$ from the built-in VAE and DiT layers, as $c_{redux}$ does not fully capture the detailed characteristics of the reference image. This is evidenced by the variations in the output subject details, as shown in Figure \ref{fig:redux}.

\subsection{Cross-Modality Prior Alignment}
\label{sec:cross_modality_prior_alignment}
We propose a Cross-Modality Prior Alignment mechanism to transfer the text-aligned visual representation from redux tokens to the text tokens, thereby reducing the discrepancy between textual and visual priors. Specifically, we introduce a learnable token before the concept name in the prompt and a deviation extraction module to enable the token to capture the deviation between textual and visual priors. Additionally, we apply a selective cross-modal attention mask and some training strategies to further align the text and visual priors.

\noindent\textbf{Learnable Token $S_*$.} A single word embedding is sufficient for capturing unique characteristics of a concept, as demonstrated by Textual Inversion \cite{gal2023an}. Building on this idea, we introduce a learnable token $S_*$, to model the deviation between textual and visual priors.

As illustrated in Figure \ref{fig:framework}, we prepend the learnable token $S_*$ to the concept name in the prompt. The T5 text encoder is then used to process the prompt and generate $c_{text}$. We extract the tokens corresponding to $S_*$ and the concept name from $c_{text}$, denoted as $c_{concept} \in \mathbb{R}^{(1+l) \times d}$, where $l$ represents the length of the concept tokens. Next, a Deviation Extraction Module (DEM) is applied to update $S_*$, producing $S_*'$.

\noindent\textbf{Deviation Extraction Module.} The Deviation Extraction Module (DEM) extracts the deviation between textual and visual priors to update the learnable token $S_*$. Given the concept tokens $c_{concept}$ and redux tokens $c_{redux}$, the process begins by passing $c_{concept}$ through a residual self-attention layer:
\begin{equation}
c_{concept}' = c_{concept} + SA(c_{concept}) \in \mathbb{R}^{(1+l) \times d},
\end{equation}
where $SA$ denotes the self-attention operation. The output $c_{concept}'$ is then processed through a residual cross-attention layer, interacting with $c_{redux}$:
\begin{equation}
c_{concept}' = c_{concept}' + CA(c_{concept}', c_{redux}) \in \mathbb{R}^{(1+l) \times d},
\end{equation}
where $CA$ represents the cross-attention operation, with $c_{concept}'$ as the query and $c_{redux}$ as the key and value. Finally, $c_{concept}'$ is passed through an MLP layer to produce the final output:
\begin{equation}
c_{concept}' = c_{concept}' + MLP(c_{concept}') \in \mathbb{R}^{(1+l) \times d}.
\end{equation}
The first token of $c_{concept}'$ corresponds to the updated learnable token $S_*'$, which captures the deviation between the textual and visual priors. Replacing the original token $S_*$ with $S_*'$ yields $c_{text}'$, which incorporates an enhanced textual prior that is more closely aligned with the visual prior. We refrain from replacing all updated concept tokens, as this approach may cause concept drift and the loss of concepts, as demonstrated in Section \ref{sec:ablation_study}.

\newcommand{\cmark}{\ding{51}}%
\newcommand{\xmark}{\ding{55}}%
\begin{table*}[ht]
\caption{Quantitative result on DreamBench++. CP$\cdot$PF refers to the product of concept preservation score and prompt following score, where higher values indicate a better balance between concept preservation and prompt following. The \colorbox[HTML]{EC7FA9}{first}, \colorbox[HTML]{FFB8E0}{second}, and \colorbox[HTML]{FFEDFA}{third} highest values are highlighted. Our approach achieves the optimal balance between CP and PF.}
\centering
\setlength{\tabcolsep}{0pt}  
\begin{tabular*}{\textwidth}{@{\extracolsep{\fill}}lcc*{5}{c}*{4}{c}@{}}
\toprule
\multirow{2}{*}{\textbf{Method}} & \multirow{2}{*}{\textbf{ZS}} & \multirow{2}{*}{\textbf{CP$\cdot$PF}}  & \multicolumn{5}{c}{\textbf{Concept Preservation}} & \multicolumn{4}{c}{\textbf{Prompt Following}} \\
\cmidrule(lr){4-8} \cmidrule(lr){9-12}
 &  &  & $\emph{Animal}$ $\uparrow$ & $\emph{Human}\uparrow$ & $\emph{Object}\uparrow$ & $\emph{Style}\uparrow$  & \textbf{Overall}$\uparrow$ & $\emph{Photorealistic}\uparrow$ & $\emph{Style}\uparrow$ & $\emph{Imaginative}\uparrow$ & \textbf{Overall}$\uparrow$ \\
\midrule
Textual Inversion \cite{gal2023an} & \xmark &   0.236 & 0.502 & 0.358 & 0.305 & 0.358 & 0.378 & 0.671 & 0.686 & 0.437 & 0.624  \\
DreamBooth \cite{ruiz2023dreambooth}          & \xmark &   0.356 & 0.640 & 0.199 & 0.488 & 0.476 & 0.494 & 0.789 & 0.775 & 0.504 & 0.721 \\
DreamBooth LoRA \cite{ruiz2023dreambooth}     & \xmark &   \colorbox[HTML]{FFB8E0}{0.517} & 0.751 & 0.311 & 0.543 & 0.718 & \colorbox[HTML]{FFB8E0}{0.598} & 0.898 & 0.895 & 0.754 & 0.865 \\
\midrule
IP-Adapter ViT-G \cite{ye2023ip}         & \cmark &   0.380 & 0.667 & 0.558 & 0.504 & 0.752 & 0.593 & 0.743 & 0.632 & 0.446 & 0.640  \\
IP-Adapter-Plus ViT-H \cite{ye2023ip}     & \cmark &   0.344 & 0.900 & 0.845 & 0.759 & 0.912 & \colorbox[HTML]{EC7FA9}{0.833} & 0.502 & 0.384 & 0.279 & 0.413 \\
BLIP-Diffusion \cite{li2023blip}      & \cmark &   0.271 & 0.673 & 0.557 & 0.469 & 0.507 & 0.547 & 0.581 & 0.510 & 0.303 & 0.495 \\
Emu2 \cite{sun2024generative}  & \cmark &   0.364 & 0.670 & 0.546 & 0.447 & 0.454 & 0.528 & 0.732 & 0.719 & 0.560 & 0.690 \\
OminiControl \cite{tan2024ominicontrol}         & \cmark &   0.435 & 0.576 & 0.217 & 0.495 & 0.324 & 0.459 & 0.979 & 0.952 & 0.876 & \colorbox[HTML]{EC7FA9}{0.947}\\
Diffusion Self-Distillation \cite{cai2024diffusion} & \cmark  & \colorbox[HTML]{FFEDFA}{0.461} & 0.580 & 0.319 & 0.591 & 0.308 & 0.514 & 0.932 & 0.879 & 0.851 & \colorbox[HTML]{FFB8E0}{0.896}\\

\textbf{Ours}   & \cmark &   \colorbox[HTML]{EC7FA9}{\textbf{0.521}} & 0.653 & 0.407 & 0.700 & 0.333 & \colorbox[HTML]{FFB8E0}{0.598} & 0.934 & 0.863 & 0.758 & \colorbox[HTML]{FFEDFA}{0.871}\\
\bottomrule
\end{tabular*}

\label{tab:quantitative_result_gpt} 
\end{table*}

\noindent\textbf{Selective Cross-Modal Attention Mask.} Relying solely on the textual prior from $c_{text}'$ is insufficient for generating fine-grained details of the reference concept. Therefore, we leverage the built-in VAE encoder and DiT layers to encode the reference image, obtaining reference image tokens $c_{ref}$. The fine-grained visual prior from $c_{ref}$ is then integrated through the multi-modal attention mechanism:
\begin{equation}
MMA([x_t; c_{text}'; c_{ref}] = \text{Softmax} \left( \frac{Q'' K''^\top}{\sqrt{d}}  + M\right) V'',
\end{equation}
where $M$ denotes a selective cross-modal attention mask. As illustrated in Figure \ref{fig:framework}, the attention mask $M$ prevents concept-irrelevant text tokens (e.g., "The", "other words") from attending to reference tokens $c_{ref}$. By applying this mask within MMA, the association between concept tokens $c_{concept}$ and reference image tokens $c_{ref}$ is further reinforced, enhancing the alignment between textual and visual priors.

Additionally, the RoPE \cite{su2024roformer} mechanism requires each token to have an assigned position index to differentiate tokens in multi-modal attention. Following the design of OminiControl \cite{tan2024ominicontrol}, we ensure that the position indices of reference tokens $c_{ref}$ do not overlap spatially with the noisy image tokens $x_t$.

\noindent\textbf{Training Strategy.} To enable $S_*$ to effectively capture deviation between textual and visual priors, we employ two training strategies: 1) random reference image dropout and 2) random concept name selection. During training, we randomly drop reference images by using a black image to obtain $c_{ref}'$, which lacks visual priors. This encourages the model to generate the target image based solely on the updated learnable token $S_*'$ and textual prior. Additionally, to enhance robustness to variations in concept names within the prompt, we randomly substitute the concept name "acoustic guitar" with its parent class "guitar" or a broader category such as "musical instruments".


\section{Experiments}
\subsection{Experimental Setup}

\noindent\textbf{Implementation details.} Our model is based on FLUX.1 DEV \cite{flux2024}, a latent rectified flow transformer model for text-to-image generation. We train the model using LoRA \cite{hu2022lora} with a rank of 16 on 1 NVIDIA A800 80GB GPUs, and 15,000 iterations. The Prodigy optimizer \cite{mishchenko2023prodigy} is used, with safeguard warmup and bias correction enabled, and a weight decay of 0.01. During inference, the flow-matching sampling is applied with 28 sampling steps.

\noindent\textbf{Datasets.} We use the Subject200K dataset \cite{tan2024ominicontrol} for training, which consists of 200,000 paired images across various categories, including clothing, furniture, vehicles, and animals. Each image pair maintains concept consistency while introducing natural variations in pose, lighting, and scene context. Although Subject200K provides detailed descriptions for each image, it does not support the random selection of concept names as described in \ref{sec:cross_modality_prior_alignment}. To overcome this limitation, we use DeepSeek-V3 \cite{deepseekai2024deepseekv3technicalreport} to replace target concept names in prompts with the special word $concept$, and generate the parent class and broader category for each target concept. This enables random substitution of $concept$ with different concept names. Additional details are provided in the supplementary material.

\begin{figure*}
    \centering
    \includegraphics[width=0.9\textwidth]{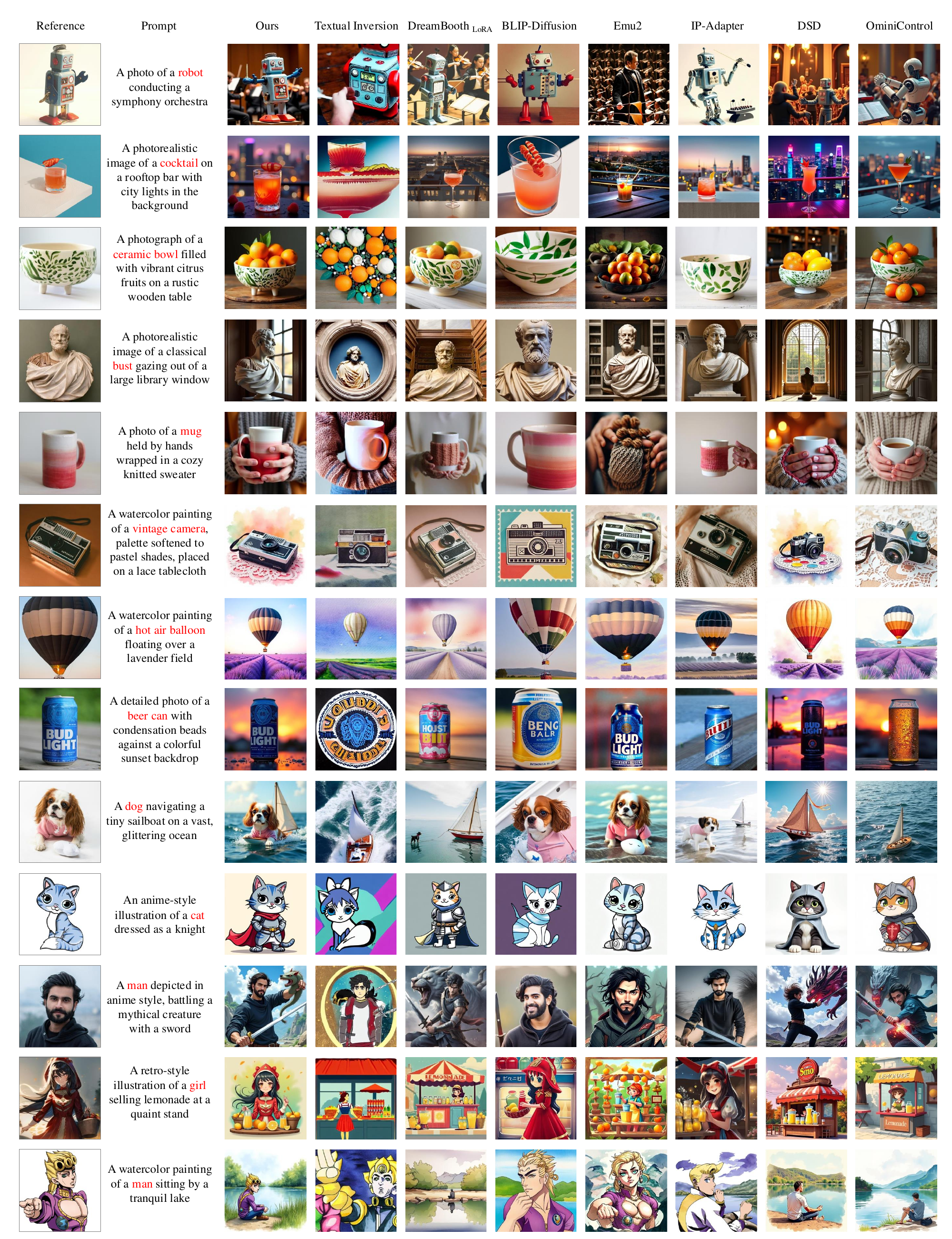}
    \caption{Qualitative comparison of the results on the Dreambench++ benchmark.}
    \Description{Qualitative comparison of the results on the Dreambench++ benchmark.}
    \label{fig:visualization_main}
\end{figure*}

\noindent\textbf{Benchmark and Evaluation Metrics.} We evaluate the performance of existing methods using DreamBench++ \cite{peng2024dreambench}, which includes 150 reference images across categories such as animals, humans, objects, and styles. Each reference image is paired with 9 prompts: 4 for photorealistic styles, 3 for non-photorealistic styles, and 2 for imaginative content, resulting in a total of 1,350 prompts.

For evaluation metrics, traditional DINO and CLIP scores are suboptimal for assessing the consistency between the generated and reference concepts, as they prioritize overall shape and color (including background elements), resulting in significant discrepancies with human preferences. In contrast, we employ GPT-4o scores, introduced in DreamBench++, to assess Concept Preservation (CP) and Prompt Following (PF). The CP measures the consistency between the generated image and the reference image, evaluating aspects such as shape, color, texture, and facial features. The PF assesses how well the generated images align with the prompt description, considering relevance, accuracy, completeness, and context. These scores are more aligned with human judgment, owing to the carefully designed evaluation instructions and advanced image understanding capabilities of GPT-4o \cite{GPT4o24}.

\noindent\textbf{Baselines.} We compare our method against two categories of approaches. The first category includes test-time optimization methods: Textual Inversion \cite{gal2023an}, DreamBooth \cite{ruiz2023dreambooth}, and DreamBooth LoRA \cite{ruiz2023dreambooth, hu2022lora}. The second category includes zero-shot methods: BLIPDiffusion \cite{li2023blip}, Emu2 \cite{sun2024generative}, IP-Adapter \cite{ye2023ip}, Diffusion Self-Distillation \cite{cai2024diffusion}, and OminiControl \cite{tan2024ominicontrol}. For methods evaluated in DreamBench++, we adopt the experimental setups in DreamBench++. For all other recent methods, including Diffusion Self-Distillation and OminiControl, we follow official implementations.

\subsection{Main Results}

\noindent\textbf{Qualitative results.} Figure \ref{fig:visualization_main} illustrates the qualitative results on the Dreambench++ benchmark, demonstrating that our method outperforms all baselines. The visualizations indicate that our approach effectively adheres to the provided prompt while preserving the key details from the reference concept, including colors, shapes, and textures. In contrast, Textual Inversion captures coarse semantics and texture from the reference image as it only uses one learnable token to learn the given concept. DreamBooth-LoRA preserves basic shapes and colors but struggles with fine-grained details, such as the dashboard on the robot or the text on the beer can. Our method leverages the powerful built-in VAE and DiT layers to encode the reference image, enabling the preservation of these fine-grained details. BLIP-Diffusion and IP-Adapter tend to reproduce the entire reference image, including the background, but struggle to adhere to the prompt due to their self-supervised learning schemes. Diffusion Self-Distillation and OminiControl fail to preserve even basic features such as color or shape in some cases (e.g., in the cocktail, mug, or hot air ballon), as these methods neglect the misalignment between textual and visual priors, leading to outputs predominantly influenced by the textual prior in the prompt. Additional visualizations in the supplementary materials further demonstrate the superiority of our method.

\begin{figure}
    \centering
    \includegraphics[width=0.45\textwidth]{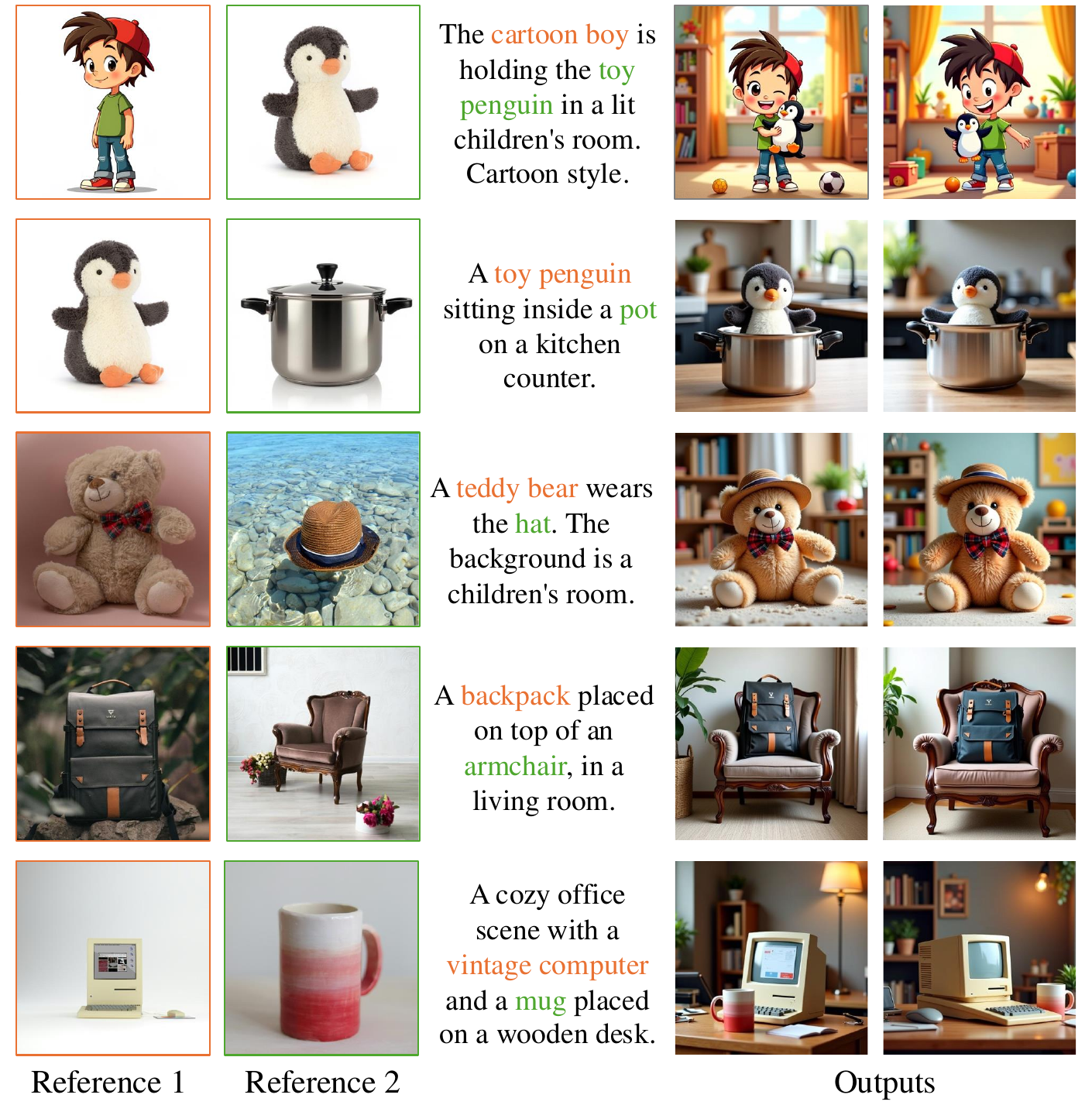}
    \caption{Qualitative results of two-subject generation without additional training. Note that the model is trained solely on single-subject datasets. The outputs correspond to different random seeds.}
    \Description{Qualitative results of two-subject generation without additional training. Note that the model is trained solely on single-subject datasets. The outputs correspond to different random seeds.}
    \label{fig:multi_ref}
\end{figure}

\noindent\textbf{Quantitative results.} 
Table \ref{tab:quantitative_result_gpt} presents the evaluation results on the DreamBench++ benchmark, including the concept preservation (CP) and prompt following (PF) scores obtained from GPT-4o. The product of CP and PF scores is the final metric as the objective is to achieve a Pareto-optimal balance between concept preservation and prompt adherence. Notably, while the IP-Adapter-Plus ViT-H model \cite{ye2023ip} demonstrates a high CP score, it struggles with prompt adherence, resulting in outputs resembling a direct copy of the reference image. Our zero-shot method achieves the best balance between concept preservation and prompt adherence, outperforming existing zero-shot approaches by 13\%, and even surpassing the test-time optimization method, DreamBooth LoRA.

\noindent\textbf{Expand to multiple reference images without additional training.} We attempt to perform personalized image generation using multiple reference images. Specifically, we prepend the same learnable token $S_*$ before each concept name in the prompt, assign the same position index to each reference image, and prevent the interactions between different reference images in the multi-modal attention. Excitingly, although our model is trained solely on single-concept datasets, it demonstrates promising generalization to multi-concept scenarios. As shown in Figure \ref{fig:multi_ref}, given two reference images, our method is able to maintain concept consistency and adhere to the prompt. However, when extended to three or more reference images, the generated results exhibit attribute confusion or omission. Despite these limitations, this finding highlights the potential of our approach for multi-concept personalization generation. In future work, we plan to construct a dedicated multi-concept training set and explore solutions to these challenges.

\subsection{Ablation Study}
In this section, we conduct ablation studies to assess the individual components of our method. Due to the time and computational cost required for performing inference on all images in the DreamBench++ and evaluating with GPT-4o, we restrict our ablation experiments to a subset of the benchmark. This subset is constructed by randomly selecting one prompt per image from the benchmark.
\label{sec:ablation_study}

\begin{figure}
    \centering
    \includegraphics[width=0.45\textwidth]{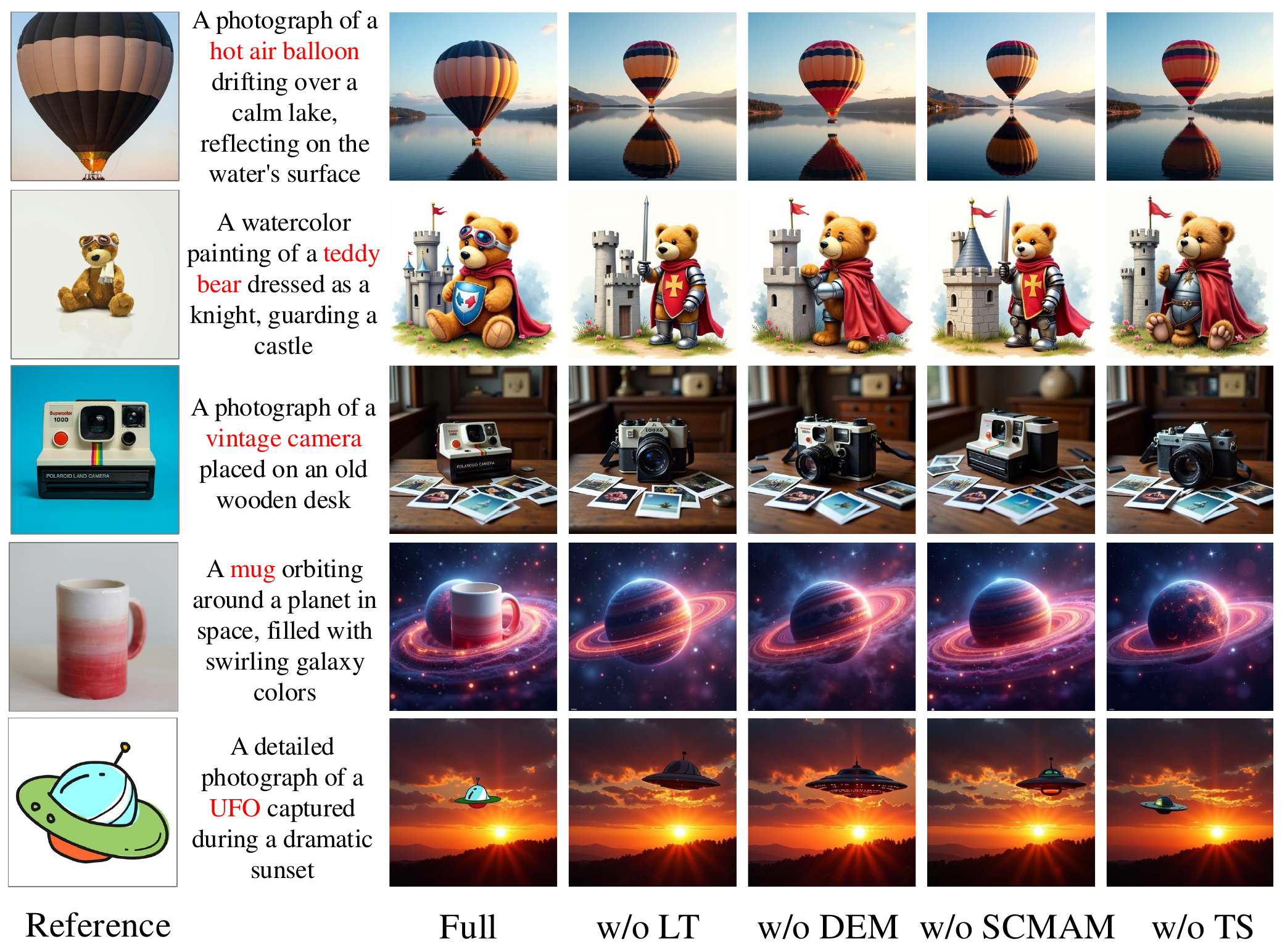}
    \caption{Visualization of inference results after removing each component. Please zoom in for a better view.}
    \Description{Visualization of inference results after removing each component. Please zoom in for a better view.}
    \label{fig:ab_components}
\end{figure}

\noindent\textbf{Effect of Different Components.} We conduct ablation studies to assess the impact of various components on performance, including the learnable token (LT), deviation extraction module (DEM), selective cross-modal attention mask (SCMAM), and training strategy (TS). As presented in Table \ref{tab:components}, the removal of individual components has a minimal effect on prompt following, but significantly affects concept preservation. The results demonstrate that omitting any component leads to a decrease in overall performance. Additionally, visualizations of the results are provided in Figure \ref{fig:ab_components}, illustrating how the removal of specific components alters the attributes of reference objects (e.g., color of the hot air balloon, style of the UFO, shape of the vintage camera), causes the disappearance of reference objects, or results in the loss of object details (e.g., glasses on the teddy bear, mug).

\begin{table}
\centering
\caption{Ablation experiments of different components. LT represents the learnable token, DEM denotes the deviation extraction module, SCMAM denotes selective cross-modal attention mask, and TS denotes training strategy. The default setting of our method is marked in gray.}
\begin{tabular}{cccc|ccc}
\toprule
 LT & DEM & SCMAM &TS & CP$\cdot$PF $\downarrow$ & CP $\downarrow$ & PF $\downarrow$  \\ \midrule
 \rowcolor[HTML]{d7dbdd}
  \ding{51} & \ding{51} & \ding{51} & \ding{51} & \textbf{0.500}          & 0.557           & 0.898 \\
  \ding{55} & \ding{51} & \ding{51} & \ding{51} & 0.463        & 0.513          & 0.903\\
  \ding{51} & \ding{55} & \ding{51} & \ding{51} & 0.451          & 0.497           & 0.908 \\
  \ding{51} & \ding{51} & \ding{55} & \ding{51} & 0.469          & 0.515           & 0.910 \\
  \ding{51} & \ding{51} & \ding{51} & \ding{55} & 0.454          & 0.502           & 0.905 \\

\bottomrule

\end{tabular}

\label{tab:components}
\end{table}
\begin{table}
\centering
\caption{Ablation experiments of different reference image drop ratio.}
\begin{tabular}{c|ccc}
\toprule
 drop ratio  & CP$\cdot$PF $\downarrow$ & CP $\downarrow$ & PF $\downarrow$  \\ \midrule
 0.1 & 0.481          & 0.540           & 0.890 \\
 0.3 & 0.492        & 0.537          & 0.917\\
 \rowcolor[HTML]{d7dbdd}
 0.5 & \textbf{0.500}          & 0.557           & 0.898 \\
 0.7 & 0.461          & 0.508           & 0.907 \\
 0.9 & 0.384          & 0.445           & 0.863 \\

\bottomrule

\end{tabular}

\label{tab:ab_drop_ratio}
\end{table}

\noindent\textbf{Effect of updating all concept tokens.} 
\begin{figure}
    \centering
    \includegraphics[width=0.45\textwidth]{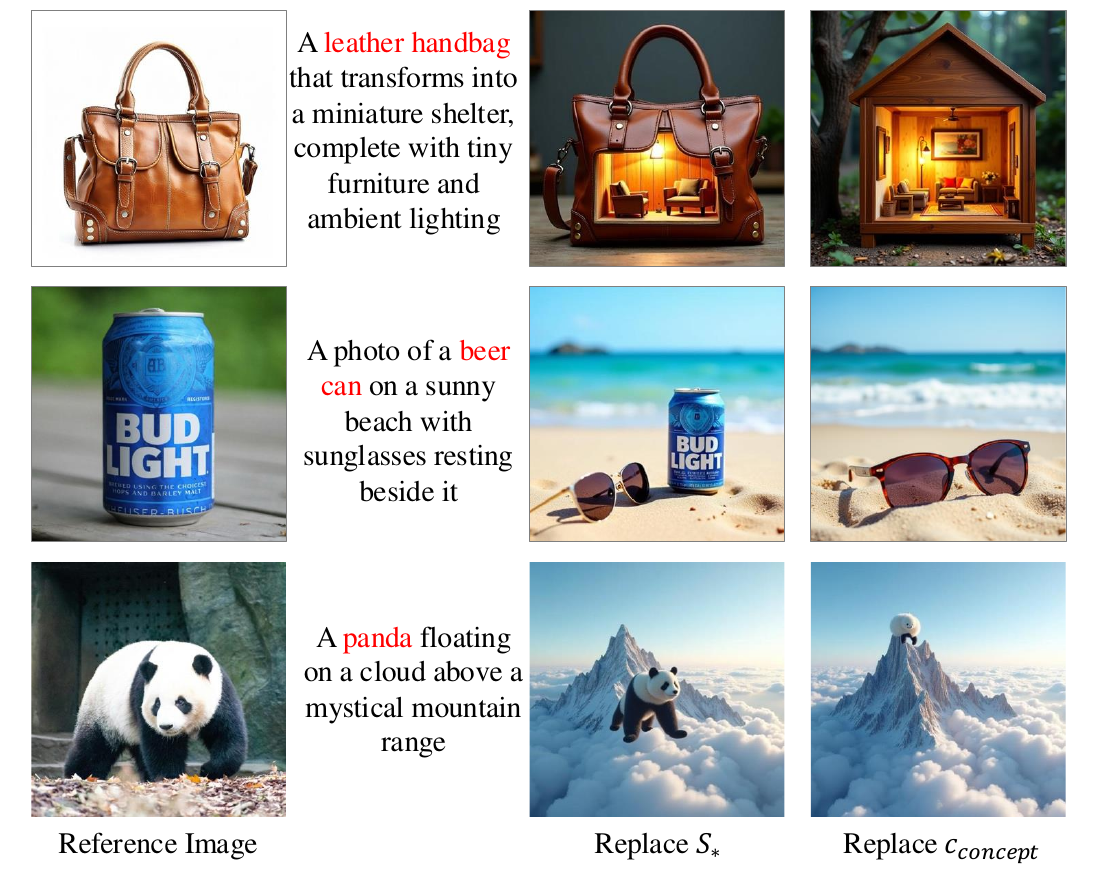}
    \caption{Effect of replacing all concept tokens.}
    \Description{Effect of replacing all concept tokens.}
    \label{fig:ab_update_all_concept_token}
\end{figure}
The deviation extraction module updates the concept tokens $c_{concept}$ to $c_{concept}'$. Instead of replacing the entire $c_{concept}$ in $c_{text}$, we only replace the first token of $c_{concept}$, corresponding to the learnable token $S_*$. The visualization results of token updates are shown in Figure \ref{fig:ab_update_all_concept_token}, which indicates that updating all tokens may induce concept drift, potentially causing the concept to disappear.

\noindent\textbf{Effect of different reference image drop ratio.} Table \ref{tab:ab_drop_ratio} shows the effects of different reference image drop ratio for training. As the drop ratio decreases from 0.1 to 0.9, the product of CP and PF decreases until reaching 0.5, beyond which it deteriorates quickly. Consequently, we select 0.5 as our default setting during training for a better trade-off between CP and PF.

\section{Conclusion and Limitation}
In this paper, we identify that the misalignment between textual prior (from the text prompt) and visual prior (from reference image) impairs the concept preservation capability in personalized image generation. To mitigate this issue, we propose a cross-modality alignment mechanism that effectively aligns the textual and visual priors. Experimental results on Dreambench++ benchmark demonstrate that our approach achieves a superior balance between concept preservation and prompt adherence compared to zero-shot and test-time optimization baselines. 

\textit{Limitation.} Although our method supports zero-shot personalized image generation for various input categories, its performance is limited when the reference image is of a human or style, due to the lack of such data in the training set. While our method shows promising generalization to multi-concept customization despite being trained only on single-concept data, it struggles when scaling to three or more reference images.



\bibliographystyle{ACM-Reference-Format}
\bibliography{sample-base}


\begin{thebibliography}{37}


\ifx \showCODEN    \undefined \def \showCODEN     #1{\unskip}     \fi
\ifx \showISBNx    \undefined \def \showISBNx     #1{\unskip}     \fi
\ifx \showISBNxiii \undefined \def \showISBNxiii  #1{\unskip}     \fi
\ifx \showISSN     \undefined \def \showISSN      #1{\unskip}     \fi
\ifx \showLCCN     \undefined \def \showLCCN      #1{\unskip}     \fi
\ifx \shownote     \undefined \def \shownote      #1{#1}          \fi
\ifx \showarticletitle \undefined \def \showarticletitle #1{#1}   \fi
\ifx \showURL      \undefined \def \showURL       {\relax}        \fi
\providecommand\bibfield[2]{#2}
\providecommand\bibinfo[2]{#2}
\providecommand\natexlab[1]{#1}
\providecommand\showeprint[2][]{arXiv:#2}

\bibitem[Alaluf et~al\mbox{.}(2023)]%
        {alaluf2023neural}
\bibfield{author}{\bibinfo{person}{Yuval Alaluf}, \bibinfo{person}{Elad Richardson}, \bibinfo{person}{Gal Metzer}, {and} \bibinfo{person}{Daniel Cohen-Or}.} \bibinfo{year}{2023}\natexlab{}.
\newblock \showarticletitle{A neural space-time representation for text-to-image personalization}.
\newblock \bibinfo{journal}{\emph{ACM Transactions on Graphics (TOG)}} \bibinfo{volume}{42}, \bibinfo{number}{6} (\bibinfo{year}{2023}), \bibinfo{pages}{1--10}.
\newblock


\bibitem[Avrahami et~al\mbox{.}(2023)]%
        {avrahami2023break}
\bibfield{author}{\bibinfo{person}{Omri Avrahami}, \bibinfo{person}{Kfir Aberman}, \bibinfo{person}{Ohad Fried}, \bibinfo{person}{Daniel Cohen-Or}, {and} \bibinfo{person}{Dani Lischinski}.} \bibinfo{year}{2023}\natexlab{}.
\newblock \showarticletitle{Break-a-scene: Extracting multiple concepts from a single image}. In \bibinfo{booktitle}{\emph{SIGGRAPH Asia 2023 Conference Papers}}. \bibinfo{pages}{1--12}.
\newblock


\bibitem[Cai et~al\mbox{.}(2024)]%
        {cai2024diffusion}
\bibfield{author}{\bibinfo{person}{Shengqu Cai}, \bibinfo{person}{Eric Chan}, \bibinfo{person}{Yunzhi Zhang}, \bibinfo{person}{Leonidas Guibas}, \bibinfo{person}{Jiajun Wu}, {and} \bibinfo{person}{Gordon Wetzstein}.} \bibinfo{year}{2024}\natexlab{}.
\newblock \showarticletitle{Diffusion self-distillation for zero-shot customized image generation}.
\newblock \bibinfo{journal}{\emph{arXiv preprint arXiv:2411.18616}} (\bibinfo{year}{2024}).
\newblock


\bibitem[DeepSeek-AI(2024)]%
        {deepseekai2024deepseekv3technicalreport}
\bibfield{author}{\bibinfo{person}{DeepSeek-AI}.} \bibinfo{year}{2024}\natexlab{}.
\newblock \bibinfo{title}{DeepSeek-V3 Technical Report}.
\newblock
\showeprint[arxiv]{2412.19437}~[cs.CL]
\urldef\tempurl%
\url{https://arxiv.org/abs/2412.19437}
\showURL{%
\tempurl}


\bibitem[Dosovitskiy et~al\mbox{.}(2021)]%
        {dosovitskiy2021an}
\bibfield{author}{\bibinfo{person}{Alexey Dosovitskiy}, \bibinfo{person}{Lucas Beyer}, \bibinfo{person}{Alexander Kolesnikov}, \bibinfo{person}{Dirk Weissenborn}, \bibinfo{person}{Xiaohua Zhai}, \bibinfo{person}{Thomas Unterthiner}, \bibinfo{person}{Mostafa Dehghani}, \bibinfo{person}{Matthias Minderer}, \bibinfo{person}{Georg Heigold}, \bibinfo{person}{Sylvain Gelly}, \bibinfo{person}{Jakob Uszkoreit}, {and} \bibinfo{person}{Neil Houlsby}.} \bibinfo{year}{2021}\natexlab{}.
\newblock \showarticletitle{An Image is Worth 16x16 Words: Transformers for Image Recognition at Scale}. In \bibinfo{booktitle}{\emph{International Conference on Learning Representations}}.
\newblock
\urldef\tempurl%
\url{https://openreview.net/forum?id=YicbFdNTTy}
\showURL{%
\tempurl}


\bibitem[Esser et~al\mbox{.}(2024)]%
        {esser2024scaling}
\bibfield{author}{\bibinfo{person}{Patrick Esser}, \bibinfo{person}{Sumith Kulal}, \bibinfo{person}{Andreas Blattmann}, \bibinfo{person}{Rahim Entezari}, \bibinfo{person}{Jonas M{\"u}ller}, \bibinfo{person}{Harry Saini}, \bibinfo{person}{Yam Levi}, \bibinfo{person}{Dominik Lorenz}, \bibinfo{person}{Axel Sauer}, \bibinfo{person}{Frederic Boesel}, \bibinfo{person}{Dustin Podell}, \bibinfo{person}{Tim Dockhorn}, \bibinfo{person}{Zion English}, {and} \bibinfo{person}{Robin Rombach}.} \bibinfo{year}{2024}\natexlab{}.
\newblock \showarticletitle{Scaling Rectified Flow Transformers for High-Resolution Image Synthesis}. In \bibinfo{booktitle}{\emph{Forty-first International Conference on Machine Learning}}.
\newblock
\urldef\tempurl%
\url{https://openreview.net/forum?id=FPnUhsQJ5B}
\showURL{%
\tempurl}


\bibitem[Gal et~al\mbox{.}(2023)]%
        {gal2023an}
\bibfield{author}{\bibinfo{person}{Rinon Gal}, \bibinfo{person}{Yuval Alaluf}, \bibinfo{person}{Yuval Atzmon}, \bibinfo{person}{Or Patashnik}, \bibinfo{person}{Amit~Haim Bermano}, \bibinfo{person}{Gal Chechik}, {and} \bibinfo{person}{Daniel Cohen-or}.} \bibinfo{year}{2023}\natexlab{}.
\newblock \showarticletitle{An Image is Worth One Word: Personalizing Text-to-Image Generation using Textual Inversion}. In \bibinfo{booktitle}{\emph{The Eleventh International Conference on Learning Representations}}.
\newblock
\urldef\tempurl%
\url{https://openreview.net/forum?id=NAQvF08TcyG}
\showURL{%
\tempurl}


\bibitem[Han et~al\mbox{.}(2023)]%
        {han2023svdiff}
\bibfield{author}{\bibinfo{person}{Ligong Han}, \bibinfo{person}{Yinxiao Li}, \bibinfo{person}{Han Zhang}, \bibinfo{person}{Peyman Milanfar}, \bibinfo{person}{Dimitris Metaxas}, {and} \bibinfo{person}{Feng Yang}.} \bibinfo{year}{2023}\natexlab{}.
\newblock \showarticletitle{Svdiff: Compact parameter space for diffusion fine-tuning}. In \bibinfo{booktitle}{\emph{Proceedings of the IEEE/CVF International Conference on Computer Vision}}. \bibinfo{pages}{7323--7334}.
\newblock


\bibitem[Ho et~al\mbox{.}(2020)]%
        {ho2020denoising}
\bibfield{author}{\bibinfo{person}{Jonathan Ho}, \bibinfo{person}{Ajay Jain}, {and} \bibinfo{person}{Pieter Abbeel}.} \bibinfo{year}{2020}\natexlab{}.
\newblock \showarticletitle{Denoising diffusion probabilistic models}.
\newblock \bibinfo{journal}{\emph{Advances in neural information processing systems}}  \bibinfo{volume}{33} (\bibinfo{year}{2020}), \bibinfo{pages}{6840--6851}.
\newblock


\bibitem[Hu et~al\mbox{.}(2022)]%
        {hu2022lora}
\bibfield{author}{\bibinfo{person}{Edward~J Hu}, \bibinfo{person}{Yelong Shen}, \bibinfo{person}{Phillip Wallis}, \bibinfo{person}{Zeyuan Allen-Zhu}, \bibinfo{person}{Yuanzhi Li}, \bibinfo{person}{Shean Wang}, \bibinfo{person}{Lu Wang}, \bibinfo{person}{Weizhu Chen}, {et~al\mbox{.}}} \bibinfo{year}{2022}\natexlab{}.
\newblock \showarticletitle{Lora: Low-rank adaptation of large language models.}
\newblock \bibinfo{journal}{\emph{ICLR}} \bibinfo{volume}{1}, \bibinfo{number}{2} (\bibinfo{year}{2022}), \bibinfo{pages}{3}.
\newblock


\bibitem[Kumari et~al\mbox{.}(2023)]%
        {kumari2023multi}
\bibfield{author}{\bibinfo{person}{Nupur Kumari}, \bibinfo{person}{Bingliang Zhang}, \bibinfo{person}{Richard Zhang}, \bibinfo{person}{Eli Shechtman}, {and} \bibinfo{person}{Jun-Yan Zhu}.} \bibinfo{year}{2023}\natexlab{}.
\newblock \showarticletitle{Multi-concept customization of text-to-image diffusion}. In \bibinfo{booktitle}{\emph{Proceedings of the IEEE/CVF conference on computer vision and pattern recognition}}. \bibinfo{pages}{1931--1941}.
\newblock


\bibitem[Labs(2024)]%
        {flux2024}
\bibfield{author}{\bibinfo{person}{Black~Forest Labs}.} \bibinfo{year}{2024}\natexlab{}.
\newblock \bibinfo{title}{FLUX}.
\newblock \bibinfo{howpublished}{\url{https://github.com/black-forest-labs/flux}}.
\newblock


\bibitem[Li et~al\mbox{.}(2023a)]%
        {li2023blip}
\bibfield{author}{\bibinfo{person}{Dongxu Li}, \bibinfo{person}{Junnan Li}, {and} \bibinfo{person}{Steven Hoi}.} \bibinfo{year}{2023}\natexlab{a}.
\newblock \showarticletitle{Blip-diffusion: Pre-trained subject representation for controllable text-to-image generation and editing}.
\newblock \bibinfo{journal}{\emph{Advances in Neural Information Processing Systems}}  \bibinfo{volume}{36} (\bibinfo{year}{2023}), \bibinfo{pages}{30146--30166}.
\newblock


\bibitem[Li et~al\mbox{.}(2023b)]%
        {li2023blip2}
\bibfield{author}{\bibinfo{person}{Junnan Li}, \bibinfo{person}{Dongxu Li}, \bibinfo{person}{Silvio Savarese}, {and} \bibinfo{person}{Steven Hoi}.} \bibinfo{year}{2023}\natexlab{b}.
\newblock \showarticletitle{Blip-2: Bootstrapping language-image pre-training with frozen image encoders and large language models}. In \bibinfo{booktitle}{\emph{International conference on machine learning}}. PMLR, \bibinfo{pages}{19730--19742}.
\newblock


\bibitem[Lipman et~al\mbox{.}(2023)]%
        {lipman2023flow}
\bibfield{author}{\bibinfo{person}{Yaron Lipman}, \bibinfo{person}{Ricky T.~Q. Chen}, \bibinfo{person}{Heli Ben-Hamu}, \bibinfo{person}{Maximilian Nickel}, {and} \bibinfo{person}{Matthew Le}.} \bibinfo{year}{2023}\natexlab{}.
\newblock \showarticletitle{Flow Matching for Generative Modeling}. In \bibinfo{booktitle}{\emph{The Eleventh International Conference on Learning Representations}}.
\newblock
\urldef\tempurl%
\url{https://openreview.net/forum?id=PqvMRDCJT9t}
\showURL{%
\tempurl}


\bibitem[Liu et~al\mbox{.}(2023)]%
        {liu2023cones}
\bibfield{author}{\bibinfo{person}{Zhiheng Liu}, \bibinfo{person}{Ruili Feng}, \bibinfo{person}{Kai Zhu}, \bibinfo{person}{Yifei Zhang}, \bibinfo{person}{Kecheng Zheng}, \bibinfo{person}{Yu Liu}, \bibinfo{person}{Deli Zhao}, \bibinfo{person}{Jingren Zhou}, {and} \bibinfo{person}{Yang Cao}.} \bibinfo{year}{2023}\natexlab{}.
\newblock \showarticletitle{Cones: Concept Neurons in Diffusion Models for Customized Generation}. In \bibinfo{booktitle}{\emph{International Conference on Machine Learning}}. PMLR, \bibinfo{pages}{21548--21566}.
\newblock


\bibitem[Ma et~al\mbox{.}(2024)]%
        {ma2024subject}
\bibfield{author}{\bibinfo{person}{Jian Ma}, \bibinfo{person}{Junhao Liang}, \bibinfo{person}{Chen Chen}, {and} \bibinfo{person}{Haonan Lu}.} \bibinfo{year}{2024}\natexlab{}.
\newblock \showarticletitle{Subject-diffusion: Open domain personalized text-to-image generation without test-time fine-tuning}. In \bibinfo{booktitle}{\emph{ACM SIGGRAPH 2024 Conference Papers}}. \bibinfo{pages}{1--12}.
\newblock


\bibitem[Mishchenko and Defazio(2023)]%
        {mishchenko2023prodigy}
\bibfield{author}{\bibinfo{person}{Konstantin Mishchenko} {and} \bibinfo{person}{Aaron Defazio}.} \bibinfo{year}{2023}\natexlab{}.
\newblock \showarticletitle{Prodigy: An expeditiously adaptive parameter-free learner}.
\newblock \bibinfo{journal}{\emph{arXiv preprint arXiv:2306.06101}} (\bibinfo{year}{2023}).
\newblock


\bibitem[Nichol et~al\mbox{.}(2022)]%
        {nichol2022glide}
\bibfield{author}{\bibinfo{person}{Alexander~Quinn Nichol}, \bibinfo{person}{Prafulla Dhariwal}, \bibinfo{person}{Aditya Ramesh}, \bibinfo{person}{Pranav Shyam}, \bibinfo{person}{Pamela Mishkin}, \bibinfo{person}{Bob Mcgrew}, \bibinfo{person}{Ilya Sutskever}, {and} \bibinfo{person}{Mark Chen}.} \bibinfo{year}{2022}\natexlab{}.
\newblock \showarticletitle{GLIDE: Towards Photorealistic Image Generation and Editing with Text-Guided Diffusion Models}. In \bibinfo{booktitle}{\emph{International Conference on Machine Learning}}. PMLR, \bibinfo{pages}{16784--16804}.
\newblock


\bibitem[OpenAI(2024)]%
        {GPT4o24}
\bibfield{author}{\bibinfo{person}{OpenAI}.} \bibinfo{year}{2024}\natexlab{}.
\newblock \showarticletitle{Introducing GPT-4o and more tools to ChatGPT free users}.
\newblock  (\bibinfo{year}{2024}).
\newblock
\urldef\tempurl%
\url{https://openai.com/index/gpt-4o-and-more-tools-to-chatgpt-free/}
\showURL{%
\tempurl}


\bibitem[Peebles and Xie(2023)]%
        {peebles2023scalable}
\bibfield{author}{\bibinfo{person}{William Peebles} {and} \bibinfo{person}{Saining Xie}.} \bibinfo{year}{2023}\natexlab{}.
\newblock \showarticletitle{Scalable diffusion models with transformers}. In \bibinfo{booktitle}{\emph{Proceedings of the IEEE/CVF international conference on computer vision}}. \bibinfo{pages}{4195--4205}.
\newblock


\bibitem[Peng et~al\mbox{.}(2025)]%
        {peng2024dreambench}
\bibfield{author}{\bibinfo{person}{Yuang Peng}, \bibinfo{person}{Yuxin Cui}, \bibinfo{person}{Haomiao Tang}, \bibinfo{person}{Zekun Qi}, \bibinfo{person}{Runpei Dong}, \bibinfo{person}{Jing Bai}, \bibinfo{person}{Chunrui Han}, \bibinfo{person}{Zheng Ge}, \bibinfo{person}{Xiangyu Zhang}, {and} \bibinfo{person}{Shu-Tao Xia}.} \bibinfo{year}{2025}\natexlab{}.
\newblock \showarticletitle{DreamBench++: A Human-Aligned Benchmark for Personalized Image Generation}. In \bibinfo{booktitle}{\emph{The Thirteenth International Conference on Learning Representations}}.
\newblock
\urldef\tempurl%
\url{https://dreambenchplus.github.io/}
\showURL{%
\tempurl}


\bibitem[Raffel et~al\mbox{.}(2020)]%
        {raffel2020exploring}
\bibfield{author}{\bibinfo{person}{Colin Raffel}, \bibinfo{person}{Noam Shazeer}, \bibinfo{person}{Adam Roberts}, \bibinfo{person}{Katherine Lee}, \bibinfo{person}{Sharan Narang}, \bibinfo{person}{Michael Matena}, \bibinfo{person}{Yanqi Zhou}, \bibinfo{person}{Wei Li}, {and} \bibinfo{person}{Peter~J Liu}.} \bibinfo{year}{2020}\natexlab{}.
\newblock \showarticletitle{Exploring the limits of transfer learning with a unified text-to-text transformer}.
\newblock \bibinfo{journal}{\emph{Journal of machine learning research}} \bibinfo{volume}{21}, \bibinfo{number}{140} (\bibinfo{year}{2020}), \bibinfo{pages}{1--67}.
\newblock


\bibitem[Ramesh et~al\mbox{.}(2022)]%
        {ramesh2022hierarchical}
\bibfield{author}{\bibinfo{person}{Aditya Ramesh}, \bibinfo{person}{Prafulla Dhariwal}, \bibinfo{person}{Alex Nichol}, \bibinfo{person}{Casey Chu}, {and} \bibinfo{person}{Mark Chen}.} \bibinfo{year}{2022}\natexlab{}.
\newblock \showarticletitle{Hierarchical text-conditional image generation with clip latents}.
\newblock \bibinfo{journal}{\emph{arXiv preprint arXiv:2204.06125}} \bibinfo{volume}{1}, \bibinfo{number}{2} (\bibinfo{year}{2022}), \bibinfo{pages}{3}.
\newblock


\bibitem[Rombach et~al\mbox{.}(2022)]%
        {rombach2022high}
\bibfield{author}{\bibinfo{person}{Robin Rombach}, \bibinfo{person}{Andreas Blattmann}, \bibinfo{person}{Dominik Lorenz}, \bibinfo{person}{Patrick Esser}, {and} \bibinfo{person}{Bj{\"o}rn Ommer}.} \bibinfo{year}{2022}\natexlab{}.
\newblock \showarticletitle{High-resolution image synthesis with latent diffusion models}. In \bibinfo{booktitle}{\emph{Proceedings of the IEEE/CVF conference on computer vision and pattern recognition}}. \bibinfo{pages}{10684--10695}.
\newblock


\bibitem[Ronneberger et~al\mbox{.}(2015)]%
        {ronneberger2015u}
\bibfield{author}{\bibinfo{person}{Olaf Ronneberger}, \bibinfo{person}{Philipp Fischer}, {and} \bibinfo{person}{Thomas Brox}.} \bibinfo{year}{2015}\natexlab{}.
\newblock \showarticletitle{U-net: Convolutional networks for biomedical image segmentation}. In \bibinfo{booktitle}{\emph{Medical image computing and computer-assisted intervention--MICCAI 2015: 18th international conference, Munich, Germany, October 5-9, 2015, proceedings, part III 18}}. Springer, \bibinfo{pages}{234--241}.
\newblock


\bibitem[Ruiz et~al\mbox{.}(2023)]%
        {ruiz2023dreambooth}
\bibfield{author}{\bibinfo{person}{Nataniel Ruiz}, \bibinfo{person}{Yuanzhen Li}, \bibinfo{person}{Varun Jampani}, \bibinfo{person}{Yael Pritch}, \bibinfo{person}{Michael Rubinstein}, {and} \bibinfo{person}{Kfir Aberman}.} \bibinfo{year}{2023}\natexlab{}.
\newblock \showarticletitle{Dreambooth: Fine tuning text-to-image diffusion models for subject-driven generation}. In \bibinfo{booktitle}{\emph{Proceedings of the IEEE/CVF conference on computer vision and pattern recognition}}. \bibinfo{pages}{22500--22510}.
\newblock


\bibitem[Saharia et~al\mbox{.}(2022)]%
        {saharia2022photorealistic}
\bibfield{author}{\bibinfo{person}{Chitwan Saharia}, \bibinfo{person}{William Chan}, \bibinfo{person}{Saurabh Saxena}, \bibinfo{person}{Lala Li}, \bibinfo{person}{Jay Whang}, \bibinfo{person}{Emily~L Denton}, \bibinfo{person}{Kamyar Ghasemipour}, \bibinfo{person}{Raphael Gontijo~Lopes}, \bibinfo{person}{Burcu Karagol~Ayan}, \bibinfo{person}{Tim Salimans}, {et~al\mbox{.}}} \bibinfo{year}{2022}\natexlab{}.
\newblock \showarticletitle{Photorealistic text-to-image diffusion models with deep language understanding}.
\newblock \bibinfo{journal}{\emph{Advances in neural information processing systems}}  \bibinfo{volume}{35} (\bibinfo{year}{2022}), \bibinfo{pages}{36479--36494}.
\newblock


\bibitem[Su et~al\mbox{.}(2024)]%
        {su2024roformer}
\bibfield{author}{\bibinfo{person}{Jianlin Su}, \bibinfo{person}{Murtadha Ahmed}, \bibinfo{person}{Yu Lu}, \bibinfo{person}{Shengfeng Pan}, \bibinfo{person}{Wen Bo}, {and} \bibinfo{person}{Yunfeng Liu}.} \bibinfo{year}{2024}\natexlab{}.
\newblock \showarticletitle{Roformer: Enhanced transformer with rotary position embedding}.
\newblock \bibinfo{journal}{\emph{Neurocomputing}}  \bibinfo{volume}{568} (\bibinfo{year}{2024}), \bibinfo{pages}{127063}.
\newblock


\bibitem[Sun et~al\mbox{.}(2024)]%
        {sun2024generative}
\bibfield{author}{\bibinfo{person}{Quan Sun}, \bibinfo{person}{Yufeng Cui}, \bibinfo{person}{Xiaosong Zhang}, \bibinfo{person}{Fan Zhang}, \bibinfo{person}{Qiying Yu}, \bibinfo{person}{Yueze Wang}, \bibinfo{person}{Yongming Rao}, \bibinfo{person}{Jingjing Liu}, \bibinfo{person}{Tiejun Huang}, {and} \bibinfo{person}{Xinlong Wang}.} \bibinfo{year}{2024}\natexlab{}.
\newblock \showarticletitle{Generative multimodal models are in-context learners}. In \bibinfo{booktitle}{\emph{Proceedings of the IEEE/CVF Conference on Computer Vision and Pattern Recognition}}. \bibinfo{pages}{14398--14409}.
\newblock


\bibitem[Tan et~al\mbox{.}(2024)]%
        {tan2024ominicontrol}
\bibfield{author}{\bibinfo{person}{Zhenxiong Tan}, \bibinfo{person}{Songhua Liu}, \bibinfo{person}{Xingyi Yang}, \bibinfo{person}{Qiaochu Xue}, {and} \bibinfo{person}{Xinchao Wang}.} \bibinfo{year}{2024}\natexlab{}.
\newblock \showarticletitle{Ominicontrol: Minimal and universal control for diffusion transformer}.
\newblock \bibinfo{journal}{\emph{arXiv preprint arXiv:2411.15098}}  \bibinfo{volume}{3} (\bibinfo{year}{2024}).
\newblock


\bibitem[Vaswani et~al\mbox{.}(2017)]%
        {vaswani2017attention}
\bibfield{author}{\bibinfo{person}{Ashish Vaswani}, \bibinfo{person}{Noam Shazeer}, \bibinfo{person}{Niki Parmar}, \bibinfo{person}{Jakob Uszkoreit}, \bibinfo{person}{Llion Jones}, \bibinfo{person}{Aidan~N Gomez}, \bibinfo{person}{{\L}ukasz Kaiser}, {and} \bibinfo{person}{Illia Polosukhin}.} \bibinfo{year}{2017}\natexlab{}.
\newblock \showarticletitle{Attention is all you need}.
\newblock \bibinfo{journal}{\emph{Advances in neural information processing systems}}  \bibinfo{volume}{30} (\bibinfo{year}{2017}).
\newblock


\bibitem[Voynov et~al\mbox{.}(2023)]%
        {voynov2023p+}
\bibfield{author}{\bibinfo{person}{Andrey Voynov}, \bibinfo{person}{Qinghao Chu}, \bibinfo{person}{Daniel Cohen-Or}, {and} \bibinfo{person}{Kfir Aberman}.} \bibinfo{year}{2023}\natexlab{}.
\newblock \showarticletitle{p+: Extended textual conditioning in text-to-image generation}.
\newblock \bibinfo{journal}{\emph{arXiv preprint arXiv:2303.09522}} (\bibinfo{year}{2023}).
\newblock


\bibitem[Wei et~al\mbox{.}(2023)]%
        {wei2023elite}
\bibfield{author}{\bibinfo{person}{Yuxiang Wei}, \bibinfo{person}{Yabo Zhang}, \bibinfo{person}{Zhilong Ji}, \bibinfo{person}{Jinfeng Bai}, \bibinfo{person}{Lei Zhang}, {and} \bibinfo{person}{Wangmeng Zuo}.} \bibinfo{year}{2023}\natexlab{}.
\newblock \showarticletitle{Elite: Encoding visual concepts into textual embeddings for customized text-to-image generation}. In \bibinfo{booktitle}{\emph{Proceedings of the IEEE/CVF International Conference on Computer Vision}}. \bibinfo{pages}{15943--15953}.
\newblock


\bibitem[Xiao et~al\mbox{.}(2024b)]%
        {xiao2024fastcomposer}
\bibfield{author}{\bibinfo{person}{Guangxuan Xiao}, \bibinfo{person}{Tianwei Yin}, \bibinfo{person}{William~T Freeman}, \bibinfo{person}{Fr{\'e}do Durand}, {and} \bibinfo{person}{Song Han}.} \bibinfo{year}{2024}\natexlab{b}.
\newblock \showarticletitle{Fastcomposer: Tuning-free multi-subject image generation with localized attention}.
\newblock \bibinfo{journal}{\emph{International Journal of Computer Vision}} (\bibinfo{year}{2024}), \bibinfo{pages}{1--20}.
\newblock


\bibitem[Xiao et~al\mbox{.}(2024a)]%
        {xiao2024omnigen}
\bibfield{author}{\bibinfo{person}{Shitao Xiao}, \bibinfo{person}{Yueze Wang}, \bibinfo{person}{Junjie Zhou}, \bibinfo{person}{Huaying Yuan}, \bibinfo{person}{Xingrun Xing}, \bibinfo{person}{Ruiran Yan}, \bibinfo{person}{Chaofan Li}, \bibinfo{person}{Shuting Wang}, \bibinfo{person}{Tiejun Huang}, {and} \bibinfo{person}{Zheng Liu}.} \bibinfo{year}{2024}\natexlab{a}.
\newblock \showarticletitle{Omnigen: Unified image generation}.
\newblock \bibinfo{journal}{\emph{arXiv preprint arXiv:2409.11340}} (\bibinfo{year}{2024}).
\newblock


\bibitem[Ye et~al\mbox{.}(2023)]%
        {ye2023ip}
\bibfield{author}{\bibinfo{person}{Hu Ye}, \bibinfo{person}{Jun Zhang}, \bibinfo{person}{Sibo Liu}, \bibinfo{person}{Xiao Han}, {and} \bibinfo{person}{Wei Yang}.} \bibinfo{year}{2023}\natexlab{}.
\newblock \showarticletitle{Ip-adapter: Text compatible image prompt adapter for text-to-image diffusion models}.
\newblock \bibinfo{journal}{\emph{arXiv preprint arXiv:2308.06721}} (\bibinfo{year}{2023}).
\newblock


\end{thebibliography}


\end{document}